# User preference extraction using dynamic query sliders in conjunction with UPS-EMO algorithm

Timo Aittokoski (`timo.aittokoski@jyu.fi`)
Suvi Tarkkanen (`suvi.p.tarkkanen@jyu.fi`)

Dept. of Mathematical Information Technology, P.O. Box 35 (Agora)
FI-40014 University of Jyväskylä, Finland

November 8, 2018

## Abstract

One drawback of evolutionary multiobjective optimization algorithms (EMOA) has traditionally been high computational cost to create an approximation of the Pareto front: number of required objective function evaluations usually grows high. On the other hand, for the decision maker (DM) it may be difficult to select one of the many produced solutions as the final one, especially in the case of more than two objectives.

To overcome the above mentioned drawbacks number of EMOA's incorporating the decision makers preference information have been proposed. In this case, it is possible to save objective function evaluations by generating only the part of the front the DM is interested in, thus also narrowing down the pool of possible selections for the final solution.

Unfortunately, most of the current EMO approaches utilizing preferences are not very intuitive to use, i.e. they may require tweaking of unintuitive parameters, and it is not always clear what kind of results one can get with given set of parameters. In this study we propose a new approach to visually inspect produced solutions, and to extract preference information from the DM to further guide the search. Our approach is based on intuitive use of dynamic query sliders, which serve as a means to extract preference information and are part of the graphical user interface implemented for the efficient UPS-EMO algorithm.

## 1 Introduction

Many industrial or engineering problems are by their very nature multiobjective ones; there are at least two conflicting objectives, meaning that one cannot be improved without impairing the another. These problems are also otherwise demanding, they cannot be solved by local methods because of several local optima. Furthermore, the solution process should be computationally efficient because objective function values may be provided by time consuming "black box" simulations. For these reasons in this study we focus on treating multi-objective, and computationally expensive global optimization problems (in box constrained domains) with a decreased budget for objective function evaluations.

A general form of a multiobjective minimization problem is

$$\begin{aligned} \text{minimize} \quad & \{f_1(\mathbf{x}), f_2(\mathbf{x}), \ldots, f_k(\mathbf{x})\} \\ \text{subject to} \quad & \mathbf{x} \in S \end{aligned} \quad (1)$$

involving $k$ ($\geq 2$) conflicting *objective functions* $f_i : \mathbb{R}^n \to \mathbb{R}$, $i = 1, ..., k$. Objective functions are minimized by altering values of the decision or design variables forming a vector $\mathbf{x} \in \mathbb{R}^n$. The *points* (or vectors) defined by values of decision variables will lie then within the search (a.k.a. decision or design) space, i.e. in box constrained domain in $\mathbb{R}^n$ in our case. Sometimes there are certain points in the search space that are not acceptable. The acceptable subset of the search space is called the *feasible region* $S$. If only box constraints are used, the search space equals to the feasible region. An *objective vector* $\mathbf{z} = \mathbf{f}(\mathbf{x}) = (f_1(\mathbf{x}), f_2(\mathbf{x}), ..., f_k(\mathbf{x}))^T$ in the *objective space* $\mathbb{R}^k$ consists of $k$ objective function values calculated in the *design (variable) vector* $\mathbf{x}$.

With multiobjective optimization (MO) problems, the concepts of non-dominated and Pareto optimal solutions are relevant. A solution (objective vector) $A$ is said to dominate solution $B$ if all components of $A$ are at least as good as those of components of $B$ (with at least one strictly better component) and $A$ is non-dominated if it is not dominated by any feasible solution. Correspondingly, solution $A$ belongs to the Pareto optimal set (in this study, the Pareto optimal set refers to a set of vectors in the objective space, and it is also referred to as Pareto front) if none of the objective function values can be improved without degrading the value of at least one objective, that is, if it is not dominated by any other feasible solution. Typically, in the presence of conflicting objectives we have many Pareto optimal solutions and we need a decision maker (DM) and his or her preference information to judge which Pareto optimal solution is the most satisfactory one as a final solution.

The aim of multiobjective optimization can be regarded to be supporting a decision maker in finding the most preferred solution to be implemented among the Pareto optimal ones [20]. In this process, the preference information specified by the decision maker is required. MO methods can be classified according to the role of the decision maker in the solution process [17, 22]. The classification given in [22] is as follows:

1. Methods where no articulation of preference information is used (no-preference methods).

2. Methods where a posteriori articulation of preference information is used (a posteriori methods).

3. Methods where a priori articulation of preference information is used (a priori methods).

4. Methods where progressive articulation of preference information is used (interactive methods).

Probably the most widely used approaches in solving demanding engineering optimization problems with multiple objectives are 1) scalarization-based methods of interactive type from the *multiple criteria decision making* (MCDM) community where the preference information is iteratively extracted from the user, and a single (or a small set of) Pareto optimal solution(s) is produced at a time (see, e.g., [22]) and, 2) a posteriori type evolutionary multiobjective optimization (EMO) approaches (see, e.g., [10]) producing an approximative representation of the Pareto optimal set, which is given to the DM for analysis after all the computation has been finished. Quality of the approximation is usually characterized by its distribution (the approximation should be diverse enough to cover the whole Pareto optimal set), and closeness to the actual Pareto optimal set.

Both approaches have some drawbacks. Even though only those Pareto optimal solutions are generated in interactive MCDM methods that the DM is interested in, computational complexity may blur the interactive nature of the iterative solution process if the DM must wait for long for solutions to be generated. At the beginning of such a process, specifying preference information may be difficult before the DM has gained some understanding of the problem (see, e.g., [19]), thus leading the DM possibly to wrong part of the Pareto front. Further, performance of the MCDM approach may be inferior to efficient EMO approaches, at least in certain cases, as suggested in [15].

On the other hand, a posteriori type EMO algorithms are often computationally expensive requiring lots of objective function evaluations and as the approximation usually has a high number of solutions, it may become cognitively very challenging for the DM to select the final solution(this applies not only to EMO approaches but also in general level), unless some proper analysis tool or a visualization of the set is available. Strangely, this latter part of selecting the final solution is usually completely neglected in the literature discussing a posteriori EMO approaches.

By incorporating the preference information in to the EMO approaches, it is possible to lessen computational burden, as only some smaller part, especially interesting for the DM, of the Pareto front is generated, instead of the whole front. Smaller portion of the front also narrows down the pool of possible selections for the final solution, thus reducing the cognitive load of the decision maker.

From our perspective, we see that from the cost-efficiency point of view, the EMO algorithm should incorporate the user preference information in an interactive way. This may seem contradictory, as we earlier pointed out that interactive nature of the MCDM approaches may be blurred due to time consuming computation of objective function values, but with regard to EMO approach our line of thought is slightly different. We assume, that the DM has (and must have) at least some preliminary understanding about the problem, otherwise he should not have been selected as the DM in the first place. From this assumption, we deduce that there are occasions when the DM can readily give some reasonable limits wherein the objective function values must lie to be acceptable, and this information is used to guide the EMO search from the very beginning in contrast to traditional a posteriori approach where the aim is to create the approximation of the whole Pareto front. In this way, the efficiency can be improved as only solutions in the ranges interesting to the DM are pro-duced.

In the literature, several different ways to incorporate the preference model in to the EMO algorithms are considered. Here we discuss some of them briefly, and also point out some drawbacks. We assume the perspective of the DM, i.e. in the following we concentrate to discuss how the DM is supposed to provide the preference information. We concentrate on approaches which seem more intuitive from the user perspective, i.e. the ones which do not require setting of very high number of parameters or otherwise highly demanding cognitive processing. For a more thorough discussion about the topic the reader is referred to surveys [7] and [25], and especially to a Chapter 9.3. in a recent book [8].

In some methods the DM must specify preferred trade-offs between objectives. This is approach taken for example in guided evolutionary algorithm [6], where the DM is allowed to define linear maximum and minimum trade-off functions. Obviously, it may be difficult for the DM to give proper values for preferred trade-offs beforehand of the optimization process. Thus, obtained results may not necessarily reflect accurately actual wishes of the DM.

Other type of extracting the preference information is by requiring the DM to rank objectives in to a relative order of importance. For example in [9] and [18] the DM characterizes relative importance of two objectives at a time, and from this data order for all the objectives is derived. Other way to extract information about ordering of the objectives is to require the DM to rank a small number of candidate solutions, and use this information to order the objectives [14]. According to [25], the task of extracting preference information in this way may be daunting to the DM when large number of objectives is involved. Further, it is not necessarily a clear-cut decision for the DM in which order to actually rank the objectives.

Another way for the DM to express preferences is by choosing satisfactory and unsatisfactory solutions from the set presented to the DM, by defining aspiration level for each of the objectives (thus constituting a "desirable point"), or by defining worst acceptable levels for each objective. All these means are proposed in [31]. Possibly the earliest work incorporating preference information in EMO [13] utilized goal attainment, where the DM defined preferred values for all the objectives, and respective weights. This led to the finding of a non-dominated solution which under- or over-attains the specified goals to a degree represented by the given weights.

It is interesting to notice that the use of a desirable point proposed in [31], bears close resemblance to much more recent algorithms, such as [11] and [32], which both utilize the reference point. Two latter ones are based on a reasoning that a single solution does not provide a good idea of the properties of solutions near the desired region of the front. For this reason, it is preferable to produce a number of solutions, thus allowing the DM to make better and more reliable decisions. In both approaches, which are at general level strikingly similar, it is left for the DM's responsibility to adjust some parameter value which controls the extent of the distribution of solutions near the given reference point.

Another algorithm based on a reference point is proposed in [5]. In this approach the DM has more control over the distribution of solutions near the given reference point because he can define also the spread of the distribution and the direction of the distribution. However, it is arguable whether this brings any relieve to the DM, as deciding suitable values may prove to be challenging and confusing task.

In a general level the reference point is a natural way to express preference information, as the DM deals directly with objective function values, which have intuitive, natural and clear meaning to the DM. For this reason we consider reference point based approaches the most developed ones at the moment of writing this study. However, drawback of the above mentioned reference point based approaches is that the DM is provided with some solution(s) near the given preference point, but it is left to the DM to adjust some parameter value(s) which controls the extent of the distribution of solutions near the given reference point. Unfortunately, the use of these parameters in all above mentioned algorithms is very unintuitive, and the DM sees what is the effect of chosen parameter value(s) only after all (possibly time consuming) computation is finished.

The main difference of our approach to current EMO approaches utilizing the preference information is the use of limits, lower and upper bounds, for each of the objectives. To extract these limits (preference information) from the DM we propose the use of sliders (see e.g., [2]) implemented in graphical user interface. A *slider* contains minimum and maximum values for an objective, a slider bar with a drag boxes, and its current values [1]. Sliders are utilized commonly to define exclusions or inclusions in several different applications, ranging from web market places to settings in computer software.

The approach using intuitive limits for objectives seems superior to those utilizing, for example the reference point [35]. With reference point one has to determine some radius, or some neighbourhood which is interesting around the given reference point. This might be unduly demand to the DM, whereas bounds given directly as objective function values bear clear meaning, so the DM does not need to decompose preference elicitation tasks into complex requests. The closeness of the command actions to the problem domain reduces user problem solving load and stress [1]. Each command is a comprehensible action, and even novice or non-technical users are able to work with them with little or no training. With sliders, user preferences can be altered or refined, which results that the solution generation will concentrate to this specific preferred region.

In addition to preference elicitation purposes as explained above, we also propose sliders to be used for dynamic queries in order to analyse the *solutions already generated and stored*. A GUI equipped with *dynamic queries* (DQs) is "an interface in which a continuous manual adjustment of one or more data values results responsively (e.g. well within one second) in a visual representation of some function(s) of those data values" [27]. According to [21], dynamic queries enable interactive visualization of multidimensional data to facilitate decision making. Thus, we propose that using dynamic queries in conjunction with EMO (as well as with other MO approaches) increases the performance of the DM. The DM may perform interactive visual analysis by dynamically filtering the solutions by means of dynamic query sliders enabling responsive (i.e. rapid) feedback. *DQ sliders* are linked to the main visualization to filter data, and they are widely used as direct manipulation tools to perform dynamic queries in order to extract data from database systems [21].

With DQ sliders, the query (lower and upper bounds for the objectives) used for filtering can be altered or refined progressively by continuously reformulating the query, and the resulting solutions are visualized in real-time. Several studies have shown that dynamic queries lead to performance improvements and user satisfaction ([2]). It should be emphasized, that in our approach the solution analysis phase may take place in parallel with preference elicitation (applying ranges for the optimization process) and solution generation process. The responsiveness of preference elicitation and solution generation phase depends on computational complexity (e.g. whether the solution generation requires computationally demanding operations such as time taking simulations). However, while solutions are being generated to a specific region, the DM may explore and analyse some other regions at the same time with dynamic queries.

In this section we have given some very brief introduction to multiobjective optimization, discussed current EMO approaches utilizing preference information, and introduced a new approach to preference elicitation in EMO. The rest of this paper is organized as follows. In Section 2 we present backgrounds for visual analysis tools, and discuss dynamic query sliders more in depth. In Section 3 we propose a novel approach, which combines preference information (extracted with DQ sliders) to the efficient UPS-EMO algorithm [3], and we also discuss shortly properties of the related graphical user interface. Section 4 is devoted for some numerical results where we show how the proposed approach works, and finally, in Section 5 we conclude this study.

## 2  Interactive Visual Analysis Applied to MO

Choosing the final solution among the ones generated by MO methods is often regarded cognitively burdensome because of high number of Pareto optimal or non-dominated solutions available. Maintaining information (such as different conflicting solutions and their tradeoffs) in the working memory is a prerequisite for thinking, reasoning, learning and problem solving. However, the capacity of working memory (short-term memory) is limited, meaning that the DM is able to process only small amount of information at once. Without any assisting techniques no meaningful learning can occur during exploration of the solution candidates. In order to facilitate insight gaining and decision making in context of MO methods (where there are multiple solutions among which to choose from), the DM should be provided with visual representation and related interaction techniques. This is especially vital in context of *a posteriori* methods where the solution set may contain such solutions which are not of interest of the DM at all. Also solutions sets generated by so called interactive MO methods may contain a great number of unwanted solutions, so interactive visual analysis techniques are of great help also in this context. Advantages of these interactive visual analysis techniques are presented next.

GUI controls for interactive visual analysis can be characterize as external representations of user's problem which stimulate and initiate cognitive activity acting as memory aids, they provide information perceptually without need for interpretation, and anchor and structure cognitive behavior [30]. According to [28], rapid interaction can facilitate active exploration of problems in a manner that is inconceivable with static displays (i.e. less dynamic display methods). For example, users can start to pose *What if* queries spontaneously as they work through a task. Such exploration can enormously facilitate the acquisition of qualitative insight into the problem as well as reveal direct quantitative results. Rapid interaction can be performed by means of dynamic queries (DQs). DQs are widely used e.g. for data filtering purposes. In addition to *filtering*, the following categories for different types of interaction techniques are proposed in [37]: *selecting*, *exploring*, *reconfiguring*, *encoding*, *abstracting/elaborating*, and *connecting* (see [37] for further details and examples). We (the authors) are wondering why these techniques are not applied in the research fields of EMO and MCDM.

## 2.1 Dynamic Queries (DQs)

Dynamic queries has first been introduced in [1], and these techniques are nowadays widely applied to SQL-type interface of relational database as a visual alternative to SQL queries. Several studies have demonstrated that design principles of DQs lead to dramatic user performance improvements and high levels of user satisfaction ([2]), so why not use those as preference elicitation tools within multiobjective optimization systems? We think that DQs are especially suitable for MO since they enable an intuitive and a natural way to interact with the multidimensional data to facilitate intelligent decision making (see e.g. [21]). DQ techniques allow the following features which facilitate preference elicitation and insight gaining [1], [2], [12]:

1. visual representation of query (user goals or preferences) and visible limits on the query range (e.g. ideal and nadir values or a range specified by the DM),

2. visual representation of the query results (i.e. nondominated or Pareto optimal solutions),

3. rapid filtering to reduce the result sets (i.e. nondominated or Pareto optimal solutions) which "restrict the information portrayed on the screen, thereby pruning the visual clutter and enabling the analyst to see important underlying patterns",

4. rapid and incremental control of query: progressive refinement and continuous reformulation of query (i.e. user goals or preferences),

5. immediate and continuous display of the results after every query adjustment,

6. allows novice users to begin working with little training but still provides expert users with powerful features.

Furthermore, DQs offer support coping with the flood of information, find needles in haystacks, support exploratory browsing to develop intuition, find patterns and exceptions, reduce users' anxiety, and make browsing fun. According to [21] dynamic queries facilitate rapid exploration of information by real-time visual display of both query formulation and results, which is inconceivable with static displays (e.g. instead of entering the value by keyboard, immediate updates are received by sliding the drag box with a mouse).

## 2.2 Dynamic Query Sliders

In dynamic queries the query (preference elicitation, user goals) is represented by sliders which can also be regarded as filters [2], since they are linked to the main visualization to filter data. DQ slider is a very efficient tool for dynamic queries, since the feedback is quick and natural to users (i.e. the results will be immediately shown when scrolling sliders) [21]. The major advantage of sliders is that the user has the control over the data to be explored [12]. DQ sliders consist of a label, a field indicating its current value (or range), a slider bar with a drag box(es), and a value at each end of the slider bar indicating minimum and maximum values [36] (see e.g. Figure 1).

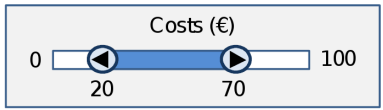

Figure 1: An example of a dynamic query slider.

According to [26], the sliders have to be placed close to the visual representation to reduce eye movement. The slider may be double-edged, containing both upper and lower thresholds [12]. Value can be changed (increased or decreased) by sliding the drag box with a mouse one step at a time, instead of entering the value by keyboard. According to [23], sliders provide a mental model of the range. For each objective a scale with adjustable sliding limits is provided. Adjustment of the upper and/or lower acceptable limits to any of these objectives causes corresponding selection to be made from all solutions, and the result of that selection is immediately displayed. [28] This way it is possible to do dozens of queries in just a few seconds [26]. There is no such thing as an "illegal" operation or incorrect syntax, so error messages are not needed. The highlighting of solutions should be in harmony with the colouring scheme of the sliders [26]. Hence, the colour area outside the specified ranges of the slider bars is the same as the solutions filtered out in the visualization side, and vice a versa. Because of this straightforward technique, even a very novice or non-technical user can use the system with little or no training.

One possible drawback of this technique is that data (solutions) are disclosed only when they satisfy the query (user preferences). There is no direct guidance as to how a query may be modified to lead to more useful information. This need is particularly acute in the situation of too many or too few hits [28]. When the data is not evenly distributed, for example, when most of the data concentrates on the lowest side, a small change on the slider can suddenly filter almost all solutions from the display, resulting in user disorientation [21]. Because of this drawback, Spence and Tweedie [28] suggest colour coding which is applied not only to those objects satisfying all objective limits but, additionally to those failing one, two, three and more such limits. For example, red colour indicates that solutions pass all limits, black indicates those that fail one limit, dark grey those that fail two limits, and so on. (See the figures in [30], and [28].) Now it is apparent how a limit may usefully be changed: if the colour of the solution is black, the user knows that one of the limits must be relaxed in order to get an acceptable solution. If the colour is grey, limits of two objectives should be relaxed, and so on. Colour indicates the reduction in the number of violated limits achieved by movement of a limit. According to [4] this approach of colour coding helps increase the flexibility of queries resulting close but inexact matches. Instead or in addition to colouring, the coding may also be realized e.g. by means of different shapes.

Example applications using DQ sliders include: HomeFinder [36, 2], FilmFinder [2], Attribute Explorer [28], Influence Explorer [30], KnowCube [16, 33].

In EMO (incorporating preferences) and interactive MO, preference elicitation can be seen as query formulation where the DM expresses which kind of solutions (s)he is interested in, and thus, we claim that DQ sliders are highly applicable also in this field. The DM can flexibly and intuitively set the query criteria by defining preferred objective ranges with sliders. The DM do not need to come up with certain discrete solution to seek for. With regard to this study, properties of the graphical user interface are discussed in the following section, where the new algorithm utilizing DQ sliders is proposed.

## 3  UPS-EMO with Preferences

The basic feature of the recently published [3] UPS-EMO algorithm is the use of a population, which has no artificial size limit (hence the name Unrestricted Population Size EMO). Instead, the population contains always all the non-dominated solutions found during the optimization process, and thus the population expands. In this way, the algorithm overcomes some problems of the current EMO approaches, such as oscillation (lack of convergence), deterioration of the population, and lack of performance.

In oscillation, some solution located very near the Pareto optimal set may be replaced by some other non-dominated solution which improves diversity, but is at the same time located much farther from the Pareto optimal set. It is shown in [3] that it depends merely on the size of the population how close to the actual Pareto optimal set the traditional EMO algorithm can eventually converge, and not on the number of objective function evaluations.

If the population gets deteriorated, then in the history of all the evaluated solutions there exist solutions which dominate the solutions in the current population. If deterioration occurs, it suggests that the algorithm has wasted some objective function evaluations, and could have actually performed better. Moreover, if a population gets heavily deteriorated, it seems intuitively plausible that also the children in the future generations will be worse than with a non-deteriorated population.

The claimed computational efficiency of the UPS-EMO algorithm is explained by three separate facts, (i) at the beginning of the solution process, the population is not forced to have some arbitrary size, and thus it does not contain dominated solutions (in the absence of non-dominated ones), which may hinder the performance, and because the population contains all the non-dominated solutions generated (ii) oscillation is not possible, and thus, (iii) the population can not deteriorate later during the solution process. This means that, in contrast to algorithms utilizing both dominance and diversity preservation in selection phase (e.g. NSGA-II), with an increasing number of objective function evaluations, the UPS-EMO algorithm either continues to converge closer to the Pareto optimal set, or to generate more solutions along the set, both being desirable features for the algorithm.

In this study we incorporate the use of the DM's preference information in to the UPS-EMO algorithm by use of dynamic query sliders discussed in Section 2.2. In this way, the DM can flexibly and intuitively set query criteria by defining appealing objective ranges with sliders, and thus avoid the need to come up with certain discrete solution to seek for. Furthermore, in contrast to some current EMO approaches (e.g. [5], [32]) utilizing the reference points, in the proposed approach there is no need for the DM to determine some radius, some neighbourhood, or some other unintuitive measure which determines the interesting region around the given reference point.

Steps of the original UPS-EMO algorithm are presented in [3] as follows:

1. Initialize the population using *minsize* random points within the given search space.

2. Evaluate the objective function values of the new points.

3. Combine the current population with the new points. Identify non-dominated solutions, and move all these to the next population. If the minimum size of population is not reached, take non-dominated solutions from the remaining points, and continue until the minimum size is reached.

4. Select randomly *burstsize* points from the current population to be used as parents. Generate one new child point for every parent point using the point generation mechanism of DE [29], [24]. In the creation of the new point, all points in the current population may participate. Points which are not inside the given search space are truncated to the border, similarly as in NSGA-II.

5. Evaluate the objective function values of the child population, and if the budget for objective function evaluations is not exhausted, go back to Step 3.

As our aim is to to utilize the DM's preference information using DQ sliders to provide preferred ranges for each of the objectives we need to complement the original algorithm slightly. It is our intention that the proposed algorithm can be used incrementally and interactively, meaning that the DM can change or refine his preferences whenever he feels that necessary. Additionally, the DM may study the solutions generated so far although the optimization is running at the background. For these reasons, the graphical user interface runs independently, but in tight co-operation with the optimization algorithm.

In the proposed approach the initial sampling is performed, and after that results are visually displayed to the DM, who can then express first time the preference information with regard to objectives. The DM can also decide how many objective function evaluations the algorithm is allowed to use before it again stops and displays the new results to the DM.

To handle the preference information (preferred ranges for objectives) provided by the DM using DQ sliders in an incremental way, we rewrite the original UPS-EMO algorithm in a following way, and reference to it as Preference UPS-EMO, or for short, PUPS-EMO:

1. Initial sampling of the search space using predefined number of points. Evaluated points are stored in to set *allPts*.

2. The current population *pop* is formed by including all the non-dominated solutions from *allPts*.

3. Publish *pop* to be visualized in GUI, and check if the optimization run must be paused, i.e. if it is the case of the first iteration and the DM is required to express preferences, the DM has requested to stop the current iteration, or the current budget for objective function evaluations is exhausted. Now the DM may decide:

    (a) if he is satisfied and wants to stop the whole optimization process

    (b) if he wants to continue the process and for how many more evaluations, i.e. define the current budget

    (c) if he wants to specify new / refine the preference information, and what are the new ranges for the objectives

4. If $size(pop) < minpopsize$, add dominated solutions from the *allPts* to the *pop* until *minpopsize* is reached.

5. Check the current preferred ranges for the objectives, and create set *prefPop* by selecting only those solutions from *pop* which reside inside given ranges.

6. If $size(prefPop) < minpopsize$, add those solutions to *prefPop* from *pop* which violate the given preferences least, until *minpopsize* is reached.

7. Select randomly *burstsize* points from *prefPop* to be used as parents. Generate set *childPop* by creating one new child point for every parent point using the trial point generation mechanism of DE [29]. In the creation of the new point, all points in the *prefPop* may participate. Points which are not inside the given search space are truncated to the border, similarly as in NSGA-II.

8. Set $allPts = allPts \cup childPop$.

9. Go back to Step 2.

It is worth mentioning that the checks against the *minpopsize* in Steps 4 and 6 are active usually only in the very beginning of the optimization run, when there are only small number of non-dominated solutions found.

In Step 6 certain number of those solutions may be added to population which violate the given preferences the least. The amount of violation is calculated as a sum over all objectives, adding absolute value of each violation of ranges to sum.

The value of *burstsize* determines directly how many new trial points are produced before they are combined to *allPts* and checked for dominance. Further, *burstsize* also defines how often new solutions are published to GUI. For these reasons, with computationally expensive problems it is useful to set *burstsize=1*, i.e. to its minimum value. Thus, both for the DM and the optimization routine there is always most up-to-date information available.

Although algorithm formulated as above is designed for incremental refinement of the preferences, as well as complete change of course during the optimization process, it is also possible to use the algorithm in a priori fashion, i.e. to express preference information only once in the beginning of the process, and let the algorithm run sufficiently long. Even if quite large preferred ranges for the objectives are used, this approach should improve the performance over the situation where no preference information is utilized at all. This behavior can be seen for example in Figure 4.

## 3.1 Graphical User Interface

We propose a novel representation technique for multiple objectives, as sketched in Figure 2. The graphical user interface (GUI) consists of rows for each objective and columns for different type of data. *Objective* column contains the names for the objectives, information whether the objective is to be minimized or maximized, and the unit information to characterize the objective values. *All solutions* column contains dot plots for each objective, where one solution vector consists of vertically aligned dots. All the solutions generated so far are represented in this column. *Filter* column contains ranges and dynamic query sliders for each objective for filtering and preference elicitation purposes. The purpose of the other data columns is explained later in this section.

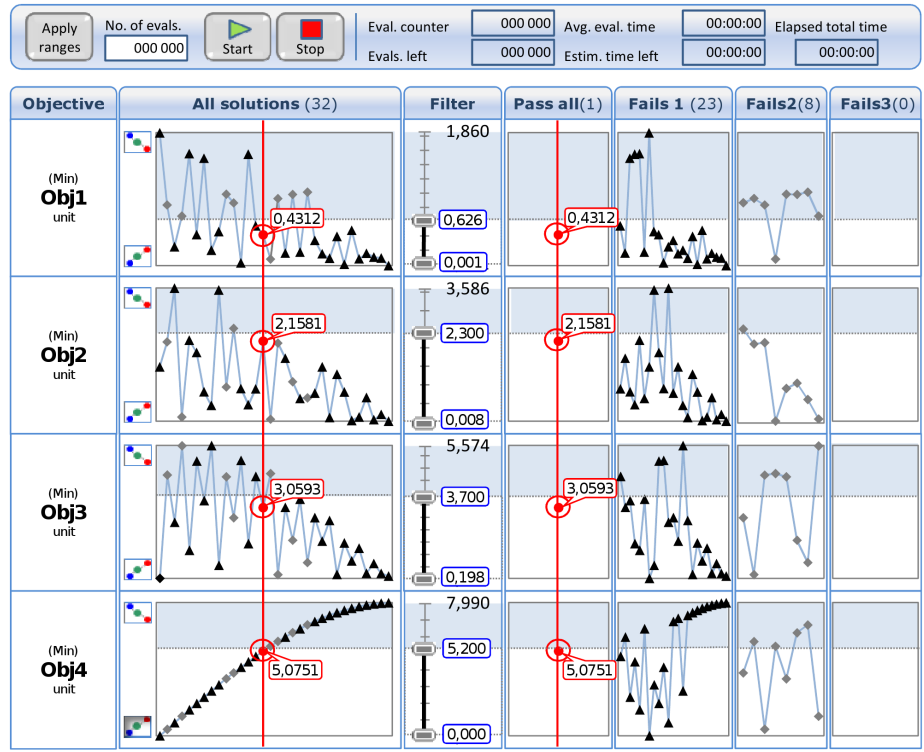

Figure 2: The graphical user interface for data analysis and preference information elicitation for PUPS-EMO.

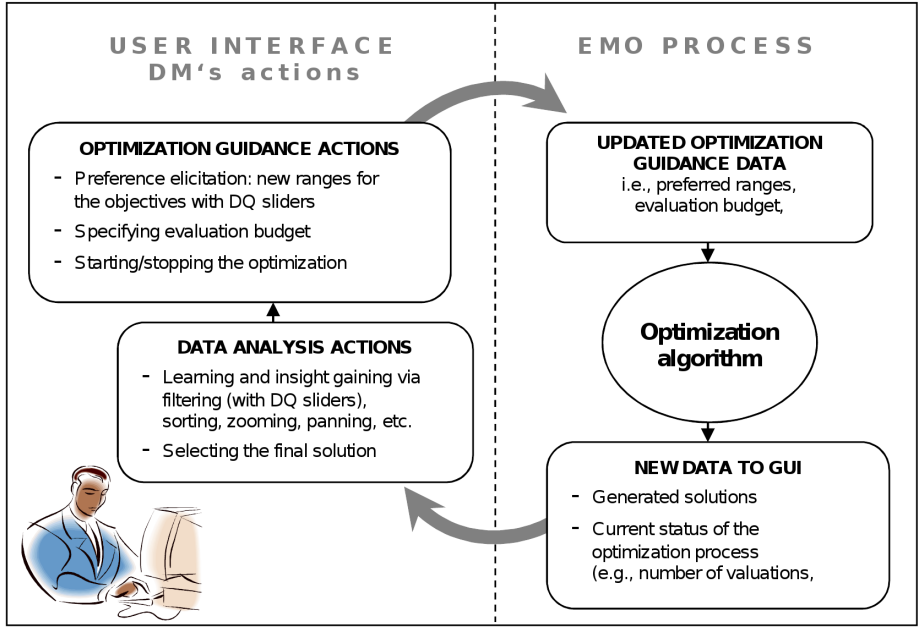

Figure 3: Iterative and interactive optimization and analysis process.

During the iterative and interactive optimization and analysis process the DM performs two kinds of actions: *optimization guidance actions* and *data analysis actions*. Thus, on the other hand, the DM guides the optimization (data generation) process by eliciting preference information, and on the other hand, he analyzes data in order to gain insights into the problem behavior (i.e. interdependencies, trade-offs, trends, correlations, etc.), and eventually to select the final solution. The DM may analyse the current solution set all the time with the tools and options provided by the GUI, even though the optimization process is running at the same time. This is possible because the GUI is implemented as a separate piece of software, and the optimization algorithm runs on its own in the background. This is obviously beneficial in the case where objective function evaluations are time consuming, for example if they are based on some black box simulations. In the description of PUPS-EMO algorithm above, the preference information is extracted at Step 3. When the DM has decided that there is need to adjust current preferences, he can immediately do so, and then inject current preferred values for the objectives in to the optimization routine, where they are applied when the routine next time passes Step 5. This behavior is better illustrated by the flowchart of the whole optimization software including both the GUI and the optimization routine, see Figure 3.

The data analysis process is facilitated by interactive visual analysis techniques, such as filtering, sorting, panning, zooming, selecting, rearranging, etc. Descriptions and examples of these interaction techniques are presented e.g. in [37]. Next, we explain how these techniques are used in our approach.

It should be emphasized that in our approach, the DQ sliders have two kinds of functionality; namely, in addition to data analysis purposes, they serve as preference elicitation tools to guide the optimization process. Let us first clarify this double role of DQ sliders.

(i) *For the data analysis*, the DQ sliders can be used as filters. By adjusting filters (see Figure 2, *Filter* column), the DM is able to change the set of solutions being presented based on some specific conditions. One may define both upper and lower limits for each of the objectives by dragging the sliders dynamically and gradually, and examine the resulting solutions of this query in real-time. According to [37], this type of interaction helps make a system feel much more responsive and live as compared to traditional batch-oriented text queries. However, in addition to defining the limits dynamically by dragging the sliders, the limits can also be entered by the keyboard via first clicking the value labels in *Filter* column with a mouse pointer.

The GUI enables advanced filtering as explained in Section 2.2: Rather than removing filtered solutions from the display, they are categorized and visualized based on how much they violate the given ranges. This helps the DM to understand the context of the solution set by showing nearby solutions not quite meeting the filtering criteria. This technique is based on the ideas presented in [30] and [28], but in our GUI we use a novel representation technique for the data to be visualized. In Figure 2 *Pass all* column contains those solutions satisfying all the objective limits, whereas *Fails 1* column contains solutions failing one such limit, *Fails 2* column containing solutions failing two limits, and so on. (In the example case there do not exist solutions failing more than two limits.)

In each column representing filtered solutions (from *Fails 1* to *Fails n*, $n$ being the number of the objectives), the solutions are sorted according to the amount of violation with respect to the given limit(s), i.e. the solutions are represented in order of superiority (computed similarly as described earlier in PUPS-EMO Step 6). Thus, "the best" nearby solutions can be found from the beginning of each column. The grouping helps the DM to notice how many limits should be readjusted in order to get the solutions from "infeasible" to "feasible area". For those solutions locating in the *Fails 1* column, only one limit has to be changed per solution vector. For example, in order to move the first solution of *Fails 1* column to *Pass all* column, the DM has to readjust one slider, that is, the upper bound for the *Obj4* above the value of 4.323. In order to move a solution from *Fails 2* to *Pass all* column, two sliders have to be readjusted: the upper limit for *Obj1* should be above 0.879 and for *Obj2* it should be above 2.499. One can see that there is only one solution passing all the given limits.

The groups (from *Fails 1* to *Fails n*) are not to be interpreted as they were representing the order of the superiority as such. In other words, the solutions locating in *Fails 1* column may be as "good" or as "bad" as solutions in *Fails n* column. The grouping only tells the DM how many bounds have to be altered (and which one(s) of them) in order to get the solutions inside the bounds. The order by which the solutions are represented *within* a group tells how close the solutions are locating with respect to the current bounds. Thus, the grouping and ordering together provide useful hints how to reformulate the query (i.e. how to readjust sliders) with a minimal effort.

(ii) *For the optimization guidance*, the DQ sliders can be used as preference elicitation tools. In case the current ranges are to be conveyed to the optimization algorithm as updated preferences, *Apply ranges* button is clicked (see Figure 2), which results that in future the data generation is to be concentrated to this specific region in consequence of PUPS-EMO algorithm. After applying the preference information, normal filtering may continue with new range adjustments independently from the optimization process, as explained before. Typically, the DM incrementally narrows down the preferred area to be applied to optimization, but it is also possible to completely change these preferences. Namely, it may happen that the ranges do not coincide at all with the Pareto front meaning that none of the data points falls on the specified region. On the other hand, the DM may be interested in several distinct areas. In case none of the solutions (or only few of them) lies within the selected range, advanced filtering technique (colour coding and grouping as explained above) will guide the DM how to modify the preferred ranges, that is, how upper and lower limits may usefully be changed. In addition to range adjustment, optimization guidance actions include defining the duration of the optimization (data generation) process by specifying evaluation budget (*No. of evals.* in Figure 2). The longer the process runs, the closer the solutions will lie to the Pareto optimal front. When the budget is exhausted, the optimization process is automatically stopped. However, it is also possible to manually stop the process by the DM (by clicking the *Stop* button). By clicking the *Start* button, optimization process springs into action again.

*All solutions* column contains sorting tools by which the solutions can be sorted by some of the objectives, ascending or descending. In Figure 2, the solutions represented in *All solutions* and *Pass all* column are sorted by *Obj4*, ascending. Sorting may reveal hidden characteristics of data, e.g. trends, correlations and interdependencies between the objectives.

For UPS-EMO it is characteristic that there may be eventually thousands of solutions generated, in which case the DM can only see a limited number of data items at a time because of screen limitations as well as limitations in human information processing. Panning and zooming can be used to explore solutions of which only part is visible at a time. In addition to defining the ranges for filtering purposes, the DM may define ranges for the solutions being visualized at a time. While panning, the DM can move on to view other solutions by grabbing the scene and moving it with a mouse resulting that new solutions enter the view as others are removed. Another way of exploring only a small set of solutions is zooming. The interesting solutions can be zoomed by defining some desired area by means of the mouse pointer. Before zooming, it may be useful to sort the solutions by the most important objective if such exists. Individual solutions can be examined more closely by dragging the vertical exploration line with a mouse pointer. When the line is locating over some individual solution vector, the solution highlighted meaning that the dots representing solution vector values are circled and labelled as presented in Figure 2. Rows and columns can be scaled both horizontally and vertically, and they can also be rearranged.

The tool box presented on the top of the Figure 2 includes some essential information related to current state of the ongoing optimization process. *Eval. counter* notifies how many evaluations are already performed, and *Evals. left* tells how many of them are still to be carried out. *Avg. eval. time* shows the average duration of an individual evaluation, *Estim. time left* tells how long it will take in wall clock time until the evaluation budget is exhausted (and the process is automatically stopped), and *Elapsed total time* expresses the time elapsed so far, after the *Start* button was clicked.

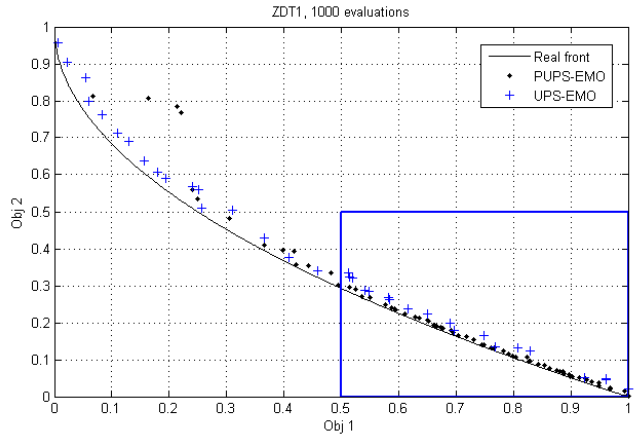

Figure 4: Comparison of the original UPS-EMO, with version utilizing preferences, PUPS-EMO, using ZDT1 function and 1000 evaluations.

## 4 Experimental Results

In this section we show how proposed algorithm PUPS-EMO performs with couple of test problems from the literature. Here we do not make systematic comparison to other algorithms or use high number of test problems; the efficiency of the original UPS-EMO is shown elsewhere [3]. Instead, we show that the use of preference information speeds up the convergence rate over the original UPS-EMO, and also that given preference ranges are adhered with high fidelity.

For PUPS-EMO algorithm, *minpopsize* was set to 10 to allow some variation in population and to prevent stagnation in very early generations, and *burstsize* to 10 because computational cost of test problems is negligible. Initial sampling was made using 100 points. For the DE [29] operators, the scaling factor $F$ was 0.8, and the crossover probability $CR$ was 0.5. The point generation strategy utilized was the one originating from a so-called *classic DE*, denoted as *DE/rand/1* strategy [24]. These DE-related values are similar to those utilized in [3].

To illustrate behavior of the proposed approach, we used two well-known test problems from the literature: bi-objective ZDT1 and ZDT3 [34]. In the Figure 4 we show that there is notable performance advantage for PUPS-EMO over UPS-EMO with ZDT1 problem after 1000 objective function evaluations. For both objectives preferred ranges were given from 0.0 to 0.5, and that region is shown as a rectangle in the figure. Solution set of PUPS-EMO is far more denser

Figure 5: Results of PUPS-EMO in two cases where the preferred ranges do not coincide with the actual Pareto front. ZDT3 test problem after 2000 evaluations.

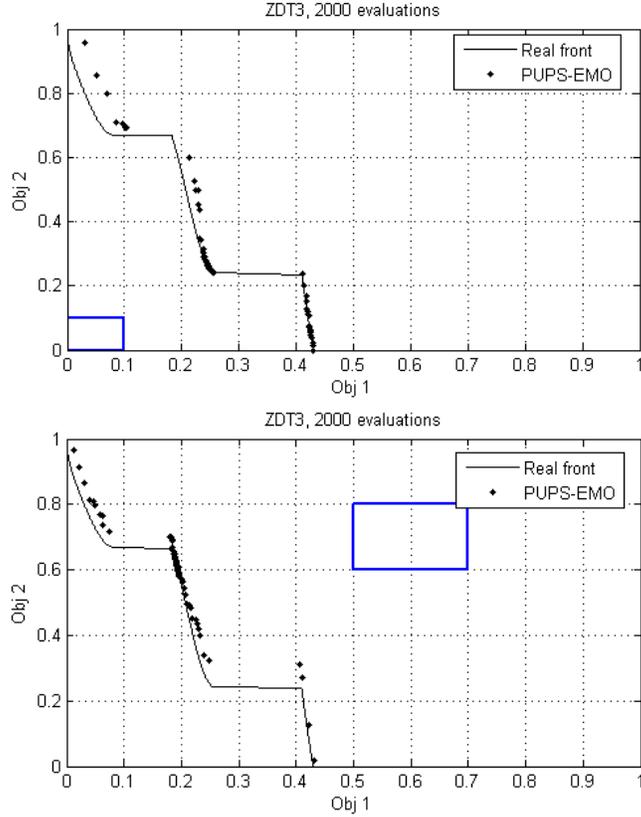

and located closer to the actual Pareto front compared to the one produced by original UPS-EMO.

In some cases the DM may not be able to define such ranges that it is possible to find solutions that satisfy his preferences. In this case the algorithm should anyhow find some solutions, that are in some sense "near" the given preferences, and can thus help to guide the search in to a proper direction. With this regard, the PUPS-EMO works reasonably well, as Figure 5 illustrates.

Next, we give an illustrative example how the DM uses the user interface connected to PUPS-EMO algorithm and equipped with DQ sliders. In all the following figures, we represent the solution set in both the traditional 2-D front representation (in a traditional way it is obviously possible to visualize at maximum 3 objectives), and also novel proposed representation utilized in GUI, which can handle more than 3 objectives.

In our example, after the initial sampling of the search space using 100 evenly distributed points, 10 solutions are found. At this point, the DM is asked for the preference information for the first time, and in this case he has more strict opinion on *Obj1* while he wants to see what kind of values *Obj2* gets when his

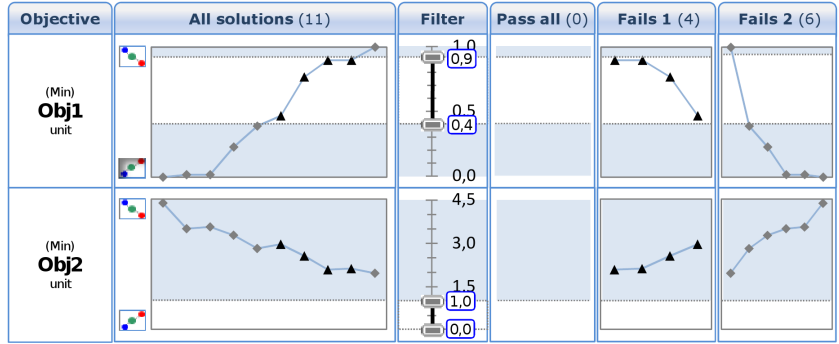
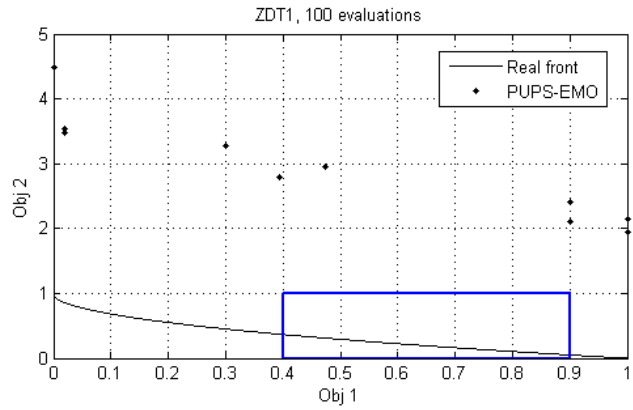

(a) Situation after 100 evaluations. Range for $Obj1 = [0.4\ 0.9]$ and for $Obj2 = [0.0\ 1.0]$.

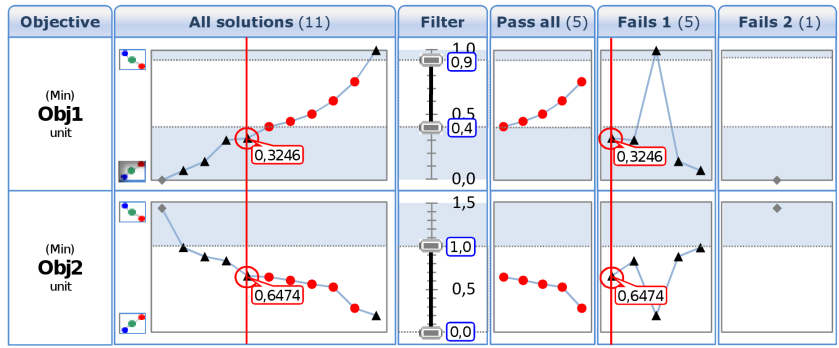
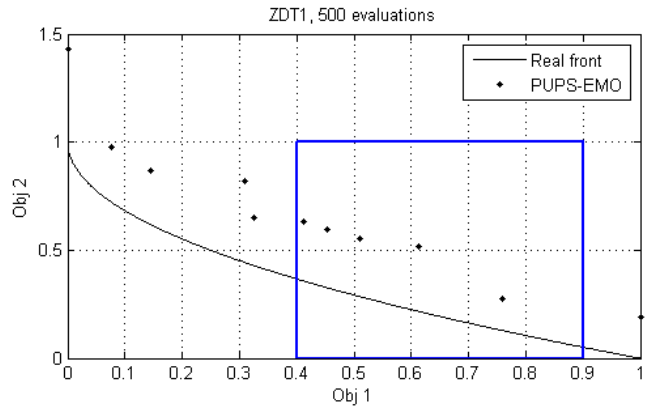

(b) Situation after 500 evaluations. Range for $Obj1 = [0.4\ 0.9]$ and for $Obj2 = [0.0\ 1.0]$.

Figure 6: Progress of PUPS-EMO run after 100 and 500 evaluations.

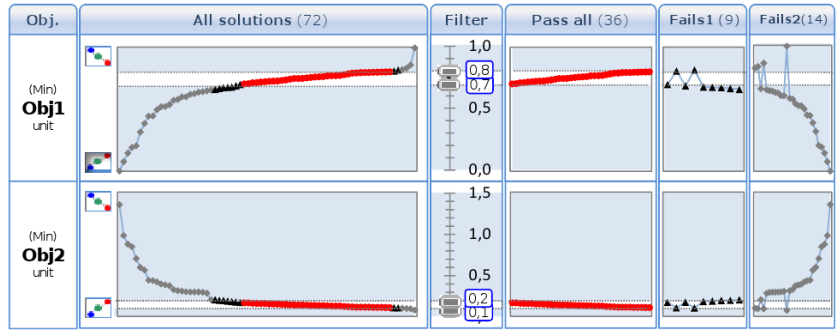

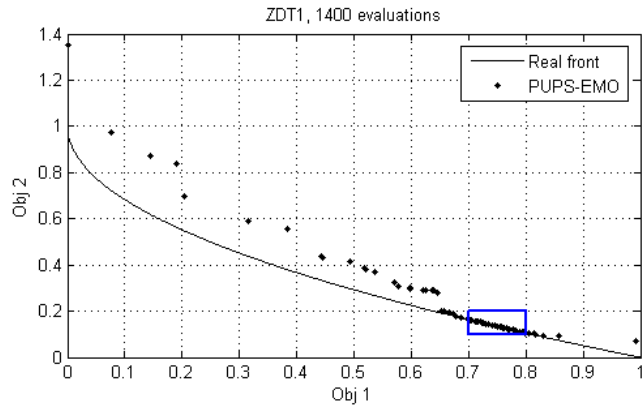

(a) Situation after 1400 evaluations. Range for $Obj1 = [0.7\ 0.8]$ and for $Obj2 = [0.1\ 0.2]$.

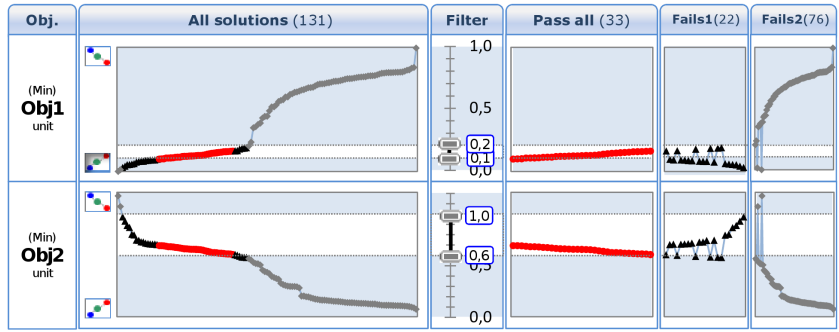

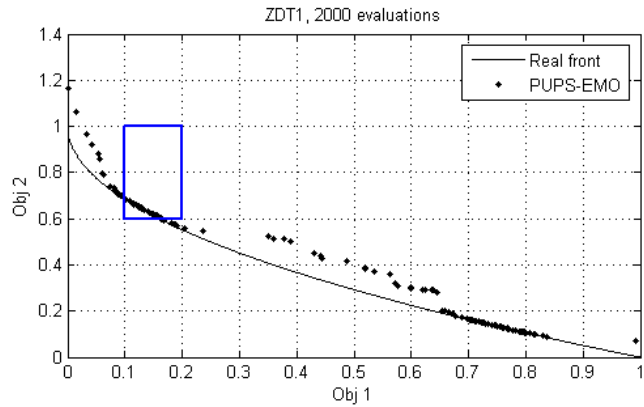

(b) Situation after 2000 evaluations. Range for $Obj1 = [0.1\ 0.2]$ and for $Obj2 = [0.6\ 1.0]$.

Figure 7: Progress of PUPS-EMO run after 1400 and 2000 evaluations.

wishes with regard to *Obj1* are satisfied. For these reasons, the DM decides to set the preferred range for *Obj1* from 0.4 to 0.9 and for *Obj2* from 0.0 to 1.0. The DM adjusts the sliders accordingly (see Figure 6(a)) and clicks the *Apply ranges* button which conveys the preference information to the PUPS-EMO algorithm. From this point on, the solution generation is concentrated to this specific region of interest.

After 400 new evaluations (500 in total) the DM stops the optimization run for a while to assess the gained results. Eleven solutions are found, five of them passing all the limits, five solutions failing one limit, and one of them failing all the limits (see Figure 6(b)).

Now the DM wishes to narrow down the ranges for *Obj1*. As he sees that there exists a solution with *Obj1* value around 0.75 and with *Obj2* value around 0.27, he decides to adjust preferred range for *Obj1* from 0.7 to 0.8 and he wants to see if he can get values for *Obj2* from 0.1 to 0.2. After 900 new evaluations (1400 in total) 72 solutions are found, 36 of them passing all the limits, see Figure 7(a).

Now, the DM is satisfied with this part of the objective space, but wants to perform *What if* analysis in the different region. For this reason, he aims his interest to other ends for both objectives and sets preferred ranges by adjusting the sliders as follows: *Obj1* should get values from 0.1 to 0.2 and *Obj2* should get values from 0.6 to 1.0. After 600 new evaluations (2000 in total) 131 solutions are found, 33 of them passing all the new limits (Figure 7(b)). At this point, the DM is satisfied with the results, and decides to end the optimization process.

From the final results of the optimization run (see Figure 7(b)) it can be clearly seen that there are two separate regions where most of the data is concentrated, exactly as the DM wished. It is noticeable that solutions are located very close to the actual Pareto optimal set, regardless of the fact that mere 2000 objective function evaluations were executed in total. Also, using the graphical user interface the nearby solutions are easily detectable. From now on, if the DM considers it necessary, the analysis process continues in a posteriori style by means of interaction and visualization techniques such as dynamic filtering, sorting, zooming, etc, to reveal the appropriate final solution set, be it a single solution or a handful of solutions.

## 5 Discussion and conclusions

In this study we have proposed a novel and very intuitive approach to extract preference information from the DM with regard to conflicting objectives. Our approach is based on dynamic query sliders, wherein the DM can visually inspect the solution set and give preferred ranges for each of the objectives as upper and lower bounds. These bounds are set using sliders, approach familiar from plethora of easy-to-use applications utilized also by laymen, such as web market places, and value settings in several different type of computer software.

As we see it, the state-of-the-art in EMO's utilizing the preference information is currently based on the use of a reference point supplied by the DM. The reference point is a natural way to express preference information, as the DM deals directly with objective function values, which have intuitive, natural and clear meaning to the DM. In current approaches the number of solutions are produced in the certain neighbourhood of the given reference point. Although this basic idea is acceptable, it seems difficult that it is left to the DM to adjust some parameter value(s) which controls the extent of the distribution of solutions near the given reference point. Unfortunately, the use of these parameters is very unintuitive, and the DM sees what is the effect of chosen parameter value(s) only after all (possibly time consuming) computation is finished.

In the proposed approach, instead of the reference point, the DM defines ranges wherein he wishes to see the solutions for each of the objectives; very similar approach is utilized in several web market places where, for example, the user looking for second hand car can set price, power, year of manufacture etc. ranges and filter away all those cars (solutions) that do not satisfy his expectations. Ranges are supplied using dynamic query sliders in the intuitive graphical user interface. In this way, vague behavior of current approaches is completely overcome, and the DM directly knows what kind of solutions to expect.

Further, the analysis and learning with regard to different solutions takes place in the graphical user interface, which runs independently but in tight cooperation with the optimization algorithm. With this arrangement, the DM can study the solutions regardless of the optimization algorithm, i.e. whether it is running or not, and if he feels the need to update the preference information, he can readily do so without interrupting the optimization algorithm, which just starts to apply the new set of preferred ranges as soon as they are supplied.

Information flow works also in to the other direction; the optimization algorithm updates always new non-dominated solutions to the graphical user interface, so that the DM can utilize the most up-to-date information in the decision making process. However, the results of a query formulation (preference elicitation with DQ sliders) for an optimization algorithm cannot be presented as dynamically as it would be possible for solutions stored readily for example in database, which is the case when analysing the data with DQ sliders. Moreover, implementing DQ sliders requires appropriate programming skills and effort.

We claim that analysing the solutions by means of dynamic query sliders enormously facilitates *What if* analysis which is an essential part of the decision making process. This kind of rapid exploration of solutions is inconceivable with static displays, but with DQ sliders the DM can start *What if* queries spontaneously as (s)he works through the multiobjective optimization problem. In pursuance of getting direct quantitative results, the DM also gains qualitative insight into the nature of the problem. Thanks to grouping supplied with the order of superiority, the DM also obtains some indication of how a query (upper/lower limits for the objectives) might usefully be modified. We also proposed a novel way of visualizing the solutions generated by an EMO method, which can handle more than three objectives. In addition, some useful interaction techniques for visual data analysis were mentioned, which would support insight gaining and eventually facilitate finding the final solution.

In the experimental part of this work we showed that in the proposed PUPS-EMO algorithm the use of preference information speeds up the convergence rate over the original UPS-EMO, and also that given preference ranges are adhered with high fidelity. We also showed that in the case where the DM is not able to define such ranges for the objectives that it is possible to find solutions that satisfy his preferences the proposed algorithm finds anyhow some solutions, that are in some sense "near" the given preferences, and can thus help to guide the search in to a proper direction.

# References

[1] C. Ahlberg, C. Williamson, and B. Shneiderman, *Dynamic queries for information exploration: An implementation and evaluation*, Proc. ACM CHI '92: Human Factors in Computing Systems, (1992) pp. 619-626.

[2] C. Ahlberg and B. Shneiderman, *Visual information seeking: tight coupling of dynamic query filters with starfield displays*. Proc. ACM CHI '94: Human Factors in Computing Systems, (1994), pp. 313-317.

[3] T. Aittokoski and K. Miettinen, *Efficient evolutionary approach to approximate the Pareto optimal set in multiobjective optimization, UPS-EMOA*, Optimization Methods and Software, 25(6)(2010), pp. 841-858.

[4] R. A. Amar and J. T. Stasko, *Knowledge precepts for design and evaluation of information visualizations*, IEEE Transactions on Visualization and Computer Graphics, 11(4)(2005), pp. 432-442.

[5] A. Auger, J. Bader, D. Brockhoff and E. Zitzler, *Articulating User Preferences in Many-Objective Problems by Sampling the Weighted Hypervolume*, In Proceedings of the 11th Annual conference on Genetic and evolutionary computation, (2009).

[6] J. Branke, T. Kaußler, H. Schmeck, *Guidance in Evolutionary Multi-Objective Optimization*, Advances in Engineering Software, 32(2001), pp. 499-507.

[7] C. A. Coello Coello, *Handling Preferences in Evolutionary Multiobjective Optimization: A Survey*, In Proceedings of the 2000 Congress on Evolutionary Computation, CEC'00, IEEE, (2000), pp. 30-37.

[8] C. A. Coello Coello, G. B. Lamont and D. A. Van Veldhuizen, *Evolutionary Algorithms for Solving Multi-Objective Problems*, 2nd Edition. Springer-Verlag, Berlin, 2007.

[9] D. Cvetkovic and I.C. Parmee, *Preferences and Their Application in Evolutionary Multiobjective Optimization*, IEEE Transactions on Evolutionary Computation, 6(1)(2002), pp. 42-57.

[10] K. Deb, *Multi-Objective Optimization using Evolutionary Algorithms*, John Wiley & Sons, Chichester, 2001.

[11] K. Deb, J. Sundar and B.R.N. Uday, *Reference Point Based Multi-Objective Optimization Using Evolutionary Algorithms*, KanGAL Report 2005012, Indian Institute of Technology. Kanpur, India, 2005.

[12] S.G. Eick, *Data visualization sliders*, Proceeding of UIST '94, ACM Press, New York, (1994), pp. 119-120.

[13] C. M. Fonseca and P. J. Fleming, *Genetic Algorithms for Multiobjective Optimization: Formulation, Discussion and Generalization*, in *Proceedings of the FiBh International Conference on Genetic Algorithms*, S. Forrest, ed., Morgan Kauffman Publishers, San Mateo, California, 1993, pp. 416-423.

[14] G.W. Greenwood, X. Hu and J.G. DAmbrosio, *Fitness functions for Multiple Objective Optimization Problems: Combining Preferences with Pareto Rankings*, In *Foundations of Genetic Algorithms 4*, R.K. Belew and M.D. Vose, eds., Morgan Kauffman Publishers, 1997, pp. 437-455.

[15] J. Hakanen and T. Aittokoski, *Comparison of MCDM and EMO Approaches in Wastewater Treatment Plant Design*, Lecture Notes in Computer Sciences (LNCS), Volume 5467, 2009, Proceedings of 5th Internatioanal Conference EMO 2009.

[16] T. Hanne and H.L. Trinkaus, *KnowCube for MCDM - Visual and interactive support for multicriteria decision making*, Published reports of the Fraunhofer ITWM, 50, Kaiserslautern, Germany, 2003.

[17] C.-L. Hwang and A. S. M. Masud, *Multiple Objective Decision Making - Methods and Applications*, Springer-Verlag, Berlin, 1979.

[18] Y. Jin and B Sendhoff, *Incorporation of Fuzzy Preferences into Evolutionary Multiobjective Optimization*, In Proceedings of the 4th Asia Pacific Conference on Simulated Evolution and Learning, 1, Singapore, 2002, pp. 26-30.

[19] K. Klamroth and K. Miettinen, *Integrating Approximation and Interactive Decision Making in Multicriteria Optimization*, Operations Research, 56(1)(2008), pp. 222-234.

[20] P. Korhonen, H. Moskowitz and J. Wallenius, *Multiple Criteria Decision Support - A Review*, European Journal of Operational Research 63(3)(1992), pp. 361-375.

[21] Q. Li and C. North, *Empirical comparison of dynamic query sliders and brushing histograms*, IEEE Symposium on Information Visualization 2003, Seattle, Washington, USA, 2003, pp. 147-153.

[22] K. Miettinen, *Nonlinear Multiobjective Optimization*, Kluwer Academic Publishers, Boston, 1999.

[23] D. A. Norman, *The Psychology of Everyday Things*, Basic Books, Inc., New York, 1988.

[24] K. V. Price, R. M. Storn and J. A. Lampinen, *Differential Evolution - A Practical Approach to Global Optimization*, Springer-Verlag, Berlin, 2005.

[25] L. Rachmawati and D. Srinivasan, *Preference Incorporation in Multi-objective Evolutionary Algorithms: A Survey*, In Proceedings of the 2006 Congress on Evolutionary Computation, CEC'06, IEEE, (2006), pp. 962-968.

[26] B. Shneiderman, C. Williamson and C. Ahlberg, *Dynamic Queries: Database Searching by Direct Manipulation*, Proc, ACM CHI, 1992.

[27] R. Spence, *Information Visualization - Design for Interaction*, 2nd Edition, Pearson Education Limited, Essex, UK, 2007, pp. 235.

[28] R. Spence and L. Tweedie, *The attribute explorer: information synthesis via exploration*, Interacting with Computers, 11 (1998), pp. 137-146.

[29] R. Storn and K. Price, *Differential Evolution - a Simple and Efficient Heuristic for Global Optimization over Continuous Spaces*, Journal of Global Optimization 11, 1997, pp. 341-359.

[30] L. Tweedie, R. Spence, H. Dawkes and H. Su, *Externalizing abstract mathematical models*. Proceedings of CHI '96, ACM, (1996), pp. 406-412.

[31] M. Tanaka and H. Watanabe, Y. Furukawa and T. Tanino, *GA-based decision support system for multicriteria optimization*, In Proceedings of IEEE International Conference on Systems, Man and Cybernetics, 2, pp. 1556-1561. Intelligent Systems for the 21st Century, New York, 1995.

[32] L. Thiele, K. Miettinen, P.J. Korhonen and J. Molina, *A Preference-Based Evolutionary Algorithm for Multi-Objective Optimization*, Evolutionary Computation, 17(3)(2009), pp. 411-436.

[33] H. L. Trinkaus and T. Hanne, *knowCube: A Visual and Interactive Support for Multicriteria Decision Making*, Computers and Operations Research, 32 (2005), pp. 1289-1309.

[34] E. Zitzler, K. Deb and L. Thiele, *Comparison of Multiobjective Evolutionary Algorithms: Empirical Results*, Evolutionary Computation, 8(2)(2000), pp. 173-195.

[35] A. P. Wierzbicki, *Reference Point Approaches*, In *Multicriteria Decision Making: Advances in MCDM Models, Algorithms, Theory, and Applications*, T. Gal, T. J. Stewart and T. Hanne, eds., Kluwer Academic Publishers, Boston, 1999.

[36] C. Williamson and B. Shneiderman, *The Dynamic HomeFinder: Evaluating dynamic queries in a real-estate information exploration system*, Proc. ACM SIGIR '92 Conference, ACM, New York, NY (1992), pp. 338-346. Reprinted in *Sparks of Innovation in Human-Computer Interaction*, B. Shneiderman, ed., Ablex Publishers, Norwood, NJ, 1993, pp. 295-307.

[37] J. S. Yi, Y. Kang, J. Stasko, J. Jacko, *Toward a deeper understanding of the role of interaction in information visualization*, IEEE Transactions on Visualization and Computer Graphics, 13(6)(2007), pp. 1224-1231.